\documentclass[letterpaper, 10pt, conference]{ieeeconf}  

\IEEEoverridecommandlockouts                              

\usepackage{amsmath} 
\usepackage{amssymb}  
\usepackage{graphicx}
\graphicspath{{images/}}
\usepackage[adjust]{cite}

\usepackage{pgfplots}
\pgfplotsset{compat=1.8}
\usepackage{tikzscale}

\usepackage{xcolor}
\usepackage{xspace}
\usepackage{csquotes}

\usepackage{algorithm}
\usepackage[noend]{algpseudocode}
\algrenewcommand\algorithmicrequire{\textbf{Given:}}
\algrenewcommand\algorithmicensure{\textbf{Wanted:}}

\usepackage{mathabx}

\usepackage{siunitx}
\DeclareSIUnit{\fps}{fps}

\usepackage{paralist}

\usepackage[hidelinks]{hyperref}

\title{\LARGE \bf
Collective perception for tracking people with a robot swarm}

\author{Miquel Kegeleirs$^{1}$, David Garzón Ramos$^{1}$, Guillermo Legarda Herranz$^{1}$, Ilyes Gharbi$^{1}$, Jeanne Szpirer$^{1}$,\\ Olivier Debeir$^{1}$, Ken Hasselmann$^{2}$, Lorenzo Garattoni$^{3}$, Gianpiero Francesca$^{3}$, and Mauro Birattari$^{1}$
\thanks{$^{1}$Université libre de Bruxelles. 
$^{2}$Royal Military Academy of Belgium. 
$^{3}$Toyota Motor Europe.
Correspondence to mauro.birattari@ulb.be.}%
\thanks{The project received funding from Belgium’s Wallonia-Brussels Federation through an ARC Advanced Project 2020 (Guaranteed by Optimization). MK, JS and MB acknowledge support from the Belgian Fonds de la Recherche Scientifique-FNRS. DGR acknowledges support from the Colombian Ministry of Science, Technology and Innovation – Minciencias.} 
\thanks{MK, DGR, GLH, IG and JS contributed equally to this work and should be recognized as co-first authors.  The research was directed by MB.}%
}

\begin{document}

\maketitle
\thispagestyle{empty}
\pagestyle{empty}

\begin{abstract}
Swarm perception refers to the ability of a robot swarm to utilize the perception capabilities of each individual robot, forming a collective understanding of the environment. 
Their distributed nature enables robot swarms to continuously monitor dynamic environments by maintaining a constant presence throughout the space.
In this study, we present a preliminary experiment on the collective tracking of people using a robot swarm. The experiment was conducted in simulation across four different office environments, with swarms of varying sizes. The robots were provided with images sampled from a dataset of real-world office environment pictures.
We measured the time distribution required for a robot to detect a person changing location and to propagate this information to increasing fractions of the swarm. 
The results indicate that robot swarms show significant promise in monitoring dynamic environments.
\end{abstract}

\section{INTRODUCTION}

\noindent A robot swarm~\cite{Ben2005sab} is a decentralized system characterized by locality of sensing and communication, self-organization, and redundancy~\cite{BraFerBirDor2013SI,DorTheTri2021PIEEE}.
Its distributed nature enables swarm perception, with rapid environmental data collection and continuous updates through peer-to-peer sharing.
In swarm robotics, research has focused on collective behaviors~\cite{TriCam2015hci-short} and decision-making~\cite{ValFerHam-etal2016AAMAS}, often emphasizing the importance of swarm perception.
Such system architecture is particularly valuable in scenarios involving people identification and tracking: swarms can share information about identified individuals,  providing continuous position estimates~\cite{KegGarLegGha-etal2024arxiv}.

We present a preliminary experiment on the collective tracking of people using a robot swarm.
We integrate existing face recognition-based identification techniques with exploration behaviors and communication protocols designed for a robot swarm.
By maintaining a presence in the whole environment, the swarm collectively constructs a dynamic representation that tracks people moving between rooms. 

\section{EXPERIMENTS}
%
We consider an office environment where four individuals perform daily activities, regularly moving from one location to another.
The robots navigate in the environment using random walk, identifying people they encounter.
For each identified person, a robot records their unique ID, location, and the timestamp of the last detection.
To maintain up-to-date information, robots exchange data whenever they meet.
We evaluate the swarm's ability to track the position of the different individuals.

Experiments were conducted in simulation using Gazebo with a standard mobile robot model equipped with LIDAR and IMU sensors.
Each robot is controlled by a separate ROS node and performs a variant of ballistic motion~\cite{KegGarBir2019taros-short,SpaKegGarBir2020bnaic-post-short}.
Each robot has perfect knowledge of its own absolute position but is unaware of the positions of other robots.

Inter-robots communication is limited to a radius of \qty{2.5}{\meter} and is simulated by ROS messages using a simple protocol:
1) each robot sends all its data to all other robots within range;
2) upon receiving data, a robot checks if the ID corresponds to the one of the persons it has already detected;
3) if the ID matches, the robot compares the timestamps and keeps the most recent record;
3') if it does not match, the new data record is appended to the robot's data.

Person identification is done using the PyTorch implementation of FaceNet~\cite{SchKalPhi2015cvpr} developed by Esler~\cite{Eslgithub}.
%
Both the fine-tuning of the identification pipeline and the experiments rely on a custom-built dataset.
We selected six locations in our laboratory (five rooms and a corridor) that differ in size, illumination, and content distribution.
Four OAK-D W cameras were placed in each location, matching their configuration on a mobile robot, and we recorded videos with four co-authors entering and leaving the location in turn.
We sampled \num{1604} frames from the videos to create the dataset.

\begin{figure}
    \centering
    \includegraphics[width=\columnwidth]{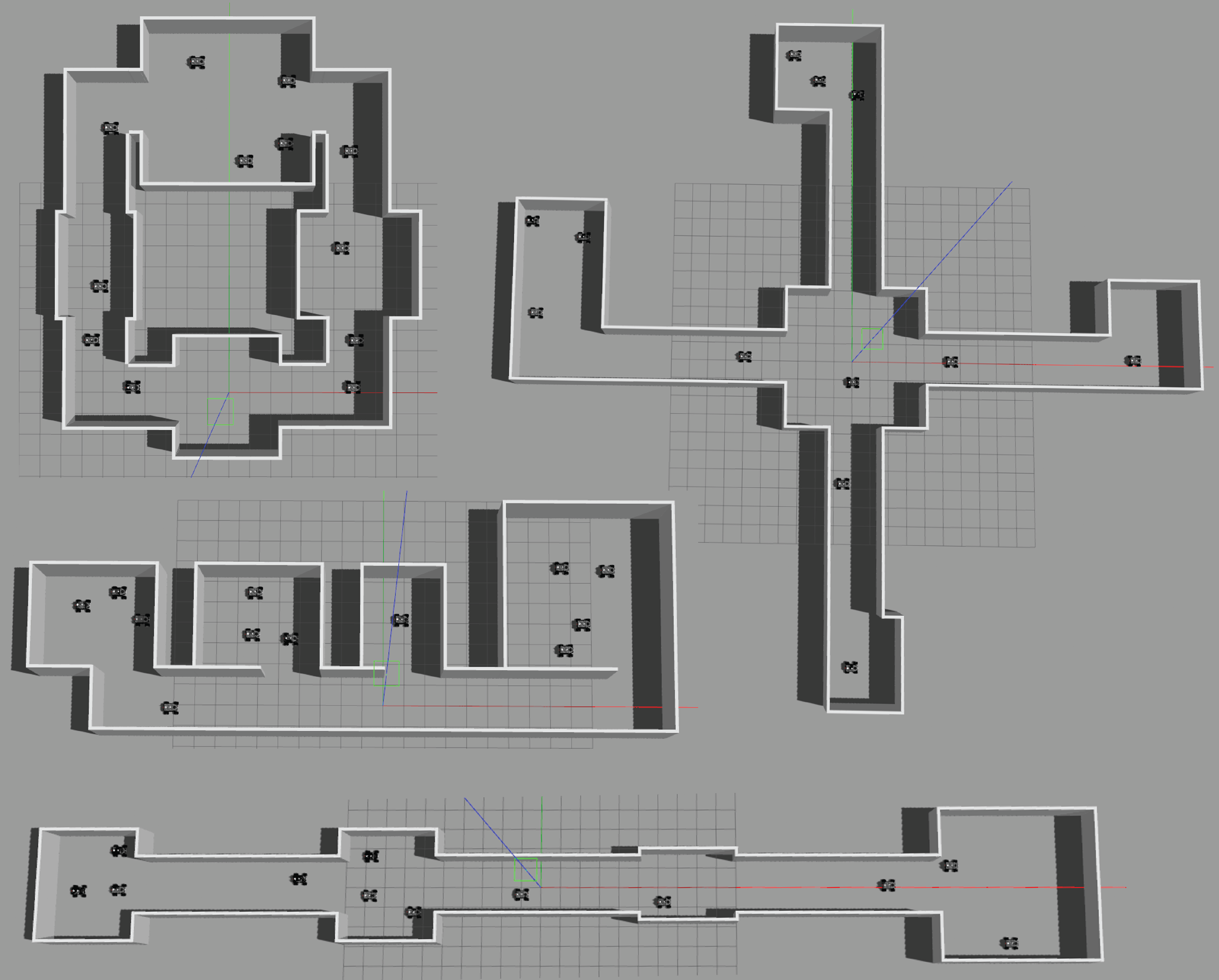}
    \caption{\label{fig:env}Simulated environments with a \num{12}-robot swarm.}
\end{figure}

In Gazebo, we recreated the five rooms and corridor, combining them to create four different environments (Fig.~\ref{fig:env}).
Automatons simulate the presence of four people in the environment.
Each automaton starts in a different location and can transition to an adjacent location every 20 seconds with a fixed probability, biased so that automatons stay longer in rooms than corridors, mimicking human behavior.
A ROS node retrieves each robot's position from the simulator and sends an image randomly sampled from a pool corresponding to i) its location and 2) present automatons.

For each environment, we performed experiments with swarms of 4, 8 and 12 robots, totaling 12 different configurations.
For each configuration, we performed 5 runs of \qty{600}{\second} (10 minutes) each.
In each run, automatons start from fixed positions (one in each room) while robots start from random positions. 
We measure the average amount of time required 1) for one of the robots to detect a person changing location (an event), and 2) for 25\%, 50\%, and 75\% of the robots to become aware of this event. 
Since people spend little time in corridors, we only consider events where a person enters a room.
Events occurring in the last \qty{180}{\second} are not considered, as there is insufficient time to propagate detections. 
The results are then aggregated across all four environments.

\begin{figure}[p!]
    \centering
    \includegraphics[width=\columnwidth]{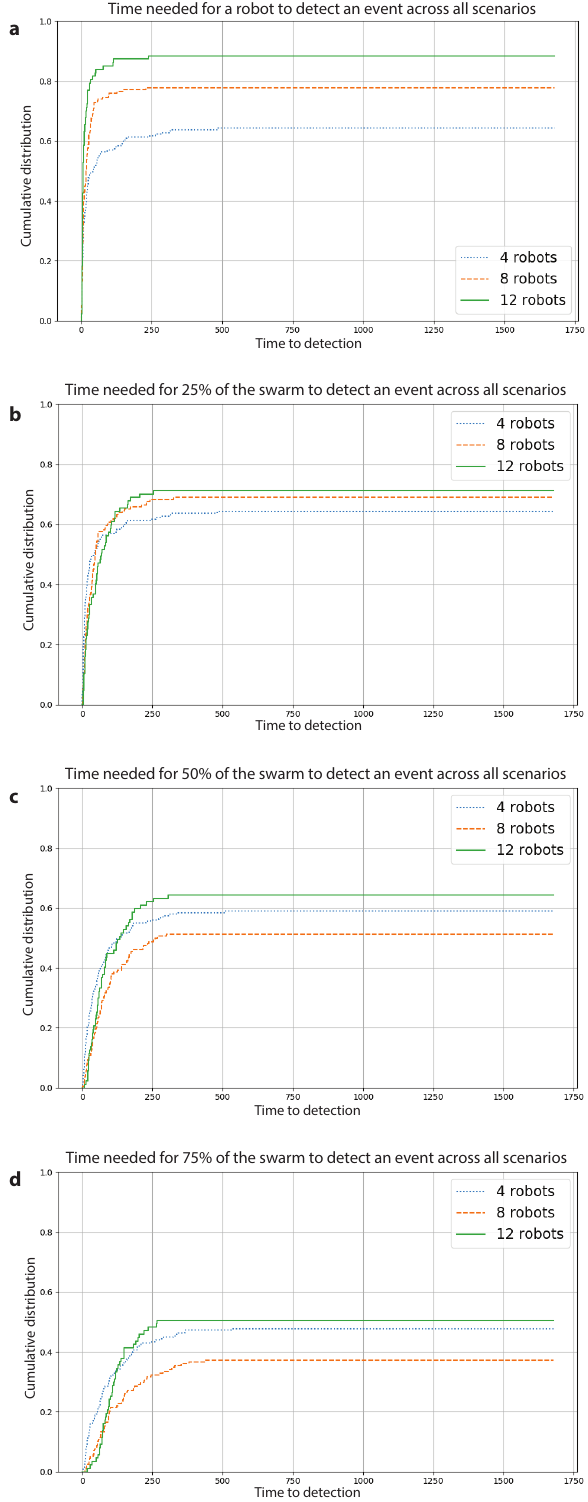}
    \caption{\label{fig:res}Empirical cumulative distribution of the time required to a) detect an event, and propagate it to b) 25\% of the swarm, c) 50\% of the swarm, and d) 75\% of the swarm, for the three different swarm sizes.}
\end{figure}

The results indicate that swarms of all sizes can detect a person's change of location more than 65\% of the time (Fig.~\ref{fig:res}-a), with the 12-robot swarm achieving a 90\% detection rate.
Most events were detected within \qty{50}{\second} after their occurrence, with larger swarms achieving shorter detection delays.
This suggests that a robot swarm appropriately sized for the environment can reliably and quickly detect events.

The first step of propagation (25\% of the swarm) is reached approximately 70\% of the time for both 8- and 12-robot swarms (Fig.~\ref{fig:res}-b).
Hence, most events detected by the swarm have sufficient time to propagate to other members before the next event occurs.
Typically, it takes \qty{250}{\second} for an event to propagate to 50\% and 75\% of the swarm (Fig.~\ref{fig:res}-c,d), which is longer than the average time for a person to move (\qty{200}{\second}).
However, the 12-robot swarm can still propagate information to 50\% of its members 65\% of the time, and to 75\% of its members more than 50\% of the time.
These results, when combined with the first analysis, suggest that the swarm can maintain sufficient cohesion for information propagation, while remaining sparse enough to cover the environment.
Robot swarms are thus promising candidates for monitoring a dynamic environment.
Additional results and resources are available in the Supplementary Materials~\cite{KegGarLeg-etal2023ICRAsupp}.%

Future work will test the proposed solution in real-robots experiments.
The hypothesis that robots know their absolute positions can be relaxed by integrating swarm SLAM algorithms~\cite{KegGriBir2021FRAI}, e.g., the one of Lajoie and Beltrame~\cite{LajBel2023IEEERAL}.
Smarter exploration behaviors could improve the results by optimizing the distribution of robots in space or the time they spend in locations.
Additional image processing could be used to improve detection or identify other useful landmarks such as doorways.
Finally, communication between the robots could be leveraged to assist the navigation.
%




%

\IEEEtriggeratref{8}
\IEEEtriggercmd{\newpage}
\bibliographystyle{IEEEtran}
\bibliography{demiurge-bib/definitions,demiurge-bib/author,demiurge-bib/address,demiurge-bib/proceedings-short,demiurge-bib/journal-short,demiurge-bib/publisher,demiurge-bib/series-short,demiurge-bib/institution,demiurge-bib/bibliography,additions}

\end{document}